\documentclass[twocolumn, switch]{article} 

\usepackage{preprint}
\usepackage{graphicx}
\usepackage{flafter}
\usepackage[colorlinks=true, urlcolor=blue, linkcolor=red]{hyperref}
\usepackage{amsmath, amsthm, amssymb, amsfonts}

\usepackage[numbers,square, sort&compress]{natbib}
\usepackage[utf8]{inputenc}	
\usepackage[T1]{fontenc}	
\usepackage{xcolor}		
\usepackage{lineno}		
\usepackage{float}
\usepackage{stfloats}			
\usepackage{caption} 
\usepackage{subcaption} 

\usepackage{lipsum}		

\usepackage{newfloat}
\DeclareFloatingEnvironment[name={Supplementary Figure}]{suppfigure}
\usepackage{sidecap}
\sidecaptionvpos{figure}{c}

\usepackage{titlesec}
\titlespacing\section{0pt}{12pt plus 3pt minus 3pt}{1pt plus 1pt minus 1pt}
\titlespacing\subsection{0pt}{10pt plus 3pt minus 3pt}{1pt plus 1pt minus 1pt}
\titlespacing\subsubsection{0pt}{8pt plus 3pt minus 3pt}{1pt plus 1pt minus 1pt}

\usepackage{tikz,xcolor,hyperref}

\definecolor{lime}{HTML}{A6CE39}
\DeclareRobustCommand{\orcidicon}{
	\begin{tikzpicture}
		\draw[lime, fill=lime] (0,0) 
		circle [radius=0.16] 
		node[white] {{\fontfamily{qag}\selectfont \tiny ID}};
		\draw[white, fill=white] (-0.0625,0.095) 
		circle [radius=0.007];
	\end{tikzpicture}
	\hspace{-2mm}
}
\foreach \x in {A, ..., Z}{\expandafter\xdef\csname orcid\x\endcsname{\noexpand\href{https://orcid.org/\csname orcidauthor\x\endcsname}
		{\noexpand\orcidicon}}
}

\title{Validation of a Zero-Shot Learning Natural Language Processing Tool for Data Abstraction from Unstructured Healthcare Data}

\usepackage[absolute,overlay]{textpos}
\usepackage{background}

\backgroundsetup{
	contents={\ifthenelse{\value{page}=1}{{\small *correspondence: \texttt{basil.kaufmann@mountsinai.org}}}{}},
	color=gray!90,
	opacity=1,
	scale=1.2,
	angle=0,
	position=current page.south,
	vshift=10mm
}

\usepackage{authblk}

\author[1, 2\thanks{\tt{basil.kaufmann@mountsinai.org}}]{Basil Kaufmann\orcidA{}}
\author[1]{Dallin Busby\orcidB{}}
\author[1]{Chandan Krushna Das\orcidC{}}
\author[1]{Neeraja Tillu\orcidF{}}
\author[1]{Mani Menon}
\author[1]{\\Ashutosh K. Tewari\orcidD{}}
\author[1]{Michael A. Gorin\orcidE{}}

\affil[1]{Milton and Carroll Petrie Department of Urology, Icahn School of Medicine at Mount Sinai, New York, USA}
\affil[2]{Department of Urology, University Hospital of Zurich, Switzerland}

\begin{document}
	\begin{textblock*}{5cm}(1cm,10cm) 
		\rotatebox{90}{\href{https://doi.org/}{\color{gray}{Publication doi}}}
	\end{textblock*}
	
	\begin{textblock*}{5cm}(20.5cm,10cm) 
		\rotatebox{90}{\href{https://doi.org/}{\color{gray}{Preprint doi}}}
		
	\end{textblock*}
	\twocolumn[
	
	\begin{@twocolumnfalse}
		
		\maketitle
		
		\begin{abstract}
			\textbf{Objectives:} To describe the development and validation of a zero-shot learning natural language processing (NLP) tool for abstracting data from unstructured text contained within PDF documents, such as those frequently found within electronic health record systems. 
			\textbf{Materials and Methods:} A data abstraction tool based on the GPT-3.5 model from OpenAI was developed and compared to three physician human abstractors in terms of time to task completion and accuracy for abstracting data on 14 unique variables from a set of 199 de-identified radical prostatectomy pathology reports. The reports were processed by the software tool in vectorized and scanned formats to establish the impact of optical character recognition on data abstraction. The tool was assessed for superiority for data abstraction speed and non-inferiority for accuracy. 
			\textbf{Results:} The human abstractors required a mean of 101 seconds (95\% CI, 97 to 104 seconds) per report for data abstraction, with times varying from 15 to 284 seconds. In comparison, the software tool required a mean of 12.8 second (95\% CI, 12.3 to 13.3 seconds) to process the vectorized reports and a mean of 15.8 seconds (95\% CI 15.1 to 16.5 seconds) to process the scanned reports (P < 0.001 for all paired comparisons).  The overall accuracies of the three human abstractors were 94.7\% (95\% CI, 93.8 to 95.5\%), 97.8\% (95\% CI, 97.2 to 98.3\%), and 96.4\% (95\% CI, 95.6 to 97\%) for the combined set of 2786 datapoints. The software tool had an overall accuracy of 94.2\% (95\% CI, 93.3 to 94.9\%) for the vectorized reports, proving to be non-inferior to the human abstractors at a margin of -10\% ($\alpha$=0.025). The tool had a slightly lower accuracy of 88.7\% (95\% CI 87.5 to 89.9\%) using the scanned reports, proving to be non-inferiority to 2 out of 3 human abstractors.
			\textbf{Conclusion:} The developed zero-shot learning NLP tool affords researchers comparable levels of accuracy to that of human abstractors, with significant time savings benefits. Because of the lack of need for task-specific model training, the developed tool is highly generalizable and can be used for a wide variety of data abstraction tasks, even outside the field of medicine.  
		\end{abstract}
		
		\keywords{data abstraction\and natural language processing\and large language models}
		\vspace{0.35cm}
		
	\end{@twocolumnfalse}
	]

	\section{Introduction}
	The advent of electronic health records (EHRs) has transformed clinical research by offering access to vast amounts of patient data. An essential part of EHRs are free-text fields. Unlike their structured counterparts, free-text fields offer a nuanced and comprehensive perspective on individual patient cases, capturing a depth of clinical information that is often not found in structured data. With that said, the use of unstructured data in clinical research is fraught with challenges\cite{RN2,RN3,RN1}. Chief among them is the labor-intensive task of data extraction. Additionally, the lack of standardization in free-text entries, due to their subjective and individualistic nature, complicates data aggregation and comparison across different records or healthcare providers. Furthermore, inconsistencies in the quality and completeness of data further heighten this challenge for researchers. This has prompted the need for sophisticated natural language processing (NLP) techniques for data abstraction, but this has the downside of introducing the potential for misinterpretation or omission of vital information\cite{RN5,RN7,RN6,RN4}. 
	
	Recent advancements in artificial intelligence technologies, particularly the introduction of large language models (LLMs) with zero-shot learning capabilities, offer a promising solution to the abstraction of unstructured healthcare data\cite{RN24, RN31, RN8, RN9, RN29}. Zero-shot learning is a concept in machine learning where a model is able to accurately classify data into categories that it has not previously encountered during its training phase. Thus, zero-shot learning makes it possible to forego the highly technical and time-consuming work of first training a model for a given data abstraction task.
	
	Herein, we describe the development of an LLM-based tool that utilizes zero-shot learning to abstract unstructured healthcare data contained within Portable Document Format (PDF) files. Following successful development of the tool, we benchmarked its performance in terms of time to task completion and accuracy for abstracting data from a set of radical prostatectomy pathology reports. We compare the performance of this tool to three physician data abstractors.

	\section{Material and Methods}
	\textit{\textbf{Software Development}}. A zero-shot learning tool for data abstraction from PDF documents was developed using Python programming language version 3.10. The completed tool incorporates functions from several third-party libraries  including PyMuPDF for text extraction \cite{RN11}, TensorFlow Hub (Google DeepMind, London, UK)  for loading a universal sentence encoder for creating semantic embedding from the extracted data\cite{RN12}, Scikit-learn for performing nearest neighbor searches\cite{RN13}, and an advanced programming interface (API) from OpenAI (San Francisco, CA) for answer generation\cite{RN14}. The software relies on the use of an API key provided by OpenAI that serves as an authentication token for accessing the “text-davinci-003” LLM. This model is based on GPT-3.5 and has been optimized for instruction-following tasks such as data abstraction\cite{RN18}.
	
\begin{figure}[h!]
	 	\centering
	 	\includegraphics[width=0.8\linewidth]{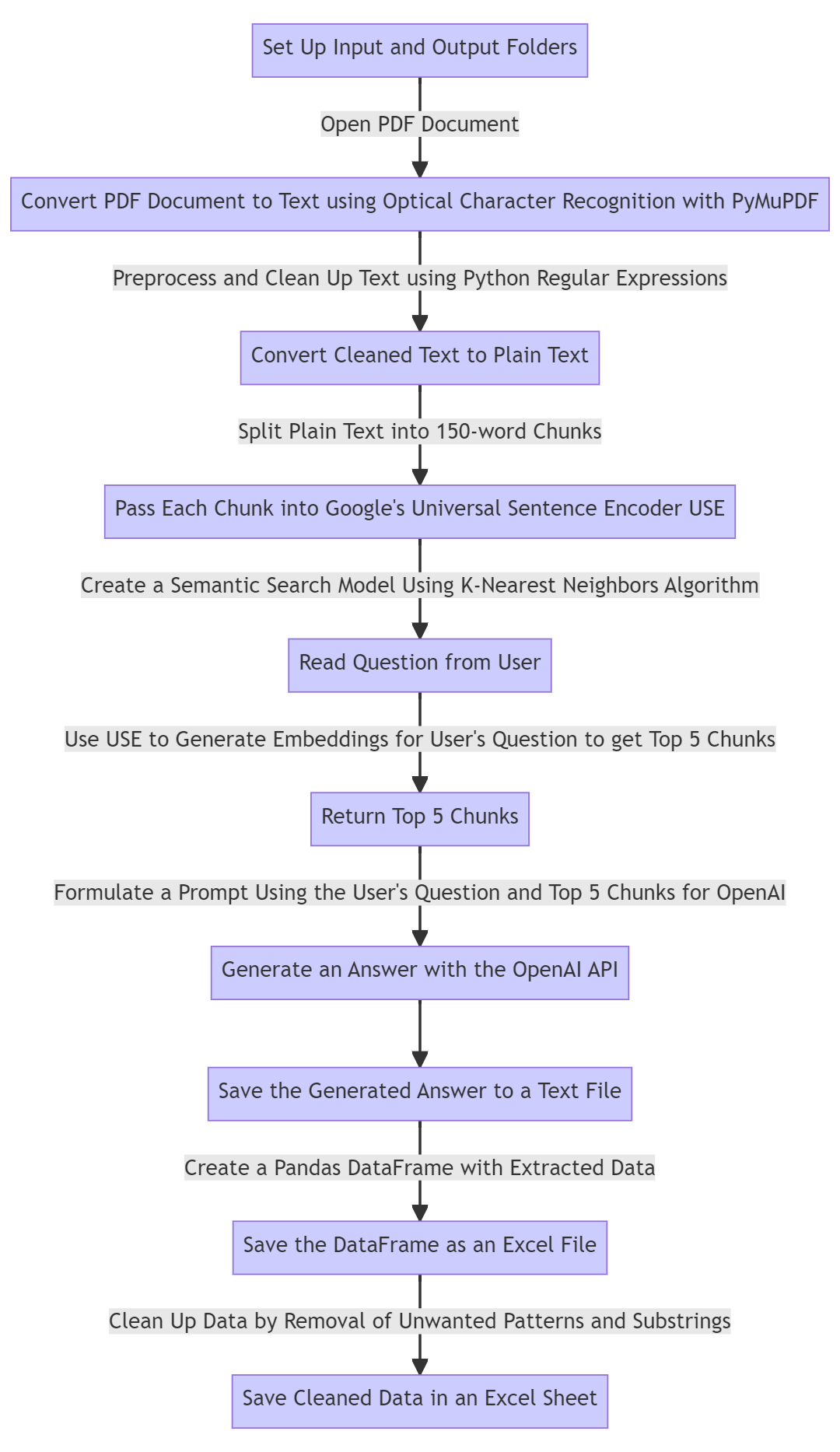}
	 	\caption{Workflow of the developed zero-shot learning NLP tool for data abstraction from unstructured text contained with PDF documents. Multiple PDF files (such as scanned pathology reports), are processed from a designated input folder and converted into text. Based on the user's question, the tool applies an AI algorithm to isolate specific information. Finally, it outputs an Excel file containing the cleaned and extracted data into a predetermined output folder.}
	 	\label{fig:fig1}
	 \end{figure}

	Figure \ref{fig:fig1} provides a graphic representation of the steps followed by the software to perform the data abstraction task. First, the program extracts text from one or more PDF files specified by the user and removes any unwanted characters or white spaces. The extracted text is next divided into smaller chunks and processed by the sentence encoder model \cite{RN12}. This step generates numerical embeddings of the text and creates an index of the data using a nearest neighbor algorithm. The software then performs a semantic search of the embedded data and finds relevant chunks of text based on the user’s input question. Next, a prompt is formulated for the text-davinci-003 model, incorporating the relevant text segments and the user’s question of interest.  The API returns the model’s response, and this text is cleaned and placed in an Excel spreadsheet (Microsoft Corporation, Redwood, WA) for downstream data cleaning, standardization, and analysis. 
	
	Because the developed tool uses zero-shot learning for data abstraction, no training steps are required for its implementation. The user is only required to provide a prompt written in plain English to query against a provided set of PDF documents. For example, a researcher may have a series of articles from which they are interested in abstracting the name of each article’s first author. The user can prompt the tool to ‘Query the input files and return the name of the first author of each article.’ The result of this will be an Excel file with a column containing author names and an identifier linking the row to its respective input PDF file. The code underlying our tool is publicly available on GitHub (\href{https://github.com/kaufmannb/PDF-Extractor}{https://github.com/kaufmannb/PDF-Extractor}). In addition, we have made available an executable file that researchers without computer coding experience can use immediately for data abstraction (\href{https://rebrand.ly/5sxu4rn}{https://rebrand.ly/5sxu4rn}). 
	
	\textit{\textbf{Datasets}}. The performance of the developed NLP tool was evaluated using a publicly available set of 199 de-identified radical prostatectomy pathology reports from The Cancer Genome Atlas (TCGA) program \cite{RN16}.  Although provided in PDF format, these pathology reports were previously printed and then scanned by the submitting sites. To mitigate potential errors introduced by optical character recognition (OCR), the reports were first converted back to a “vectorized” format as they originally appeared in the EHR. This was performed by exporting each document into .docx format using Adobe Acrobat Pro (Adobe, Inc., San Jose, CA) and then manually correcting any introduced errors in Microsoft Word. The corrected reports were saved back into PDF format, maintaining the integrity of the original formatting and layout (Supplementary Figure \ref{suppfigure:supp1}). These vectorized reports served as the input for our main round of data abstraction by the NLP tool. Additionally, a secondary analysis was performed using the original scanned reports to quantify the impact of OCR errors on the tool's performance. 
	
	\textit{\textbf{Assessment of the Data Abstraction Tool}}. Following completed development of the data abstraction tool and generation of the testing datasets, we benchmark its performance against three human data abstractors (all of whom are physicians) in terms of time to task completion and accuracy of data abstraction. The human abstractors and software tool were asked to perform the identical task of abstracting 14 variables from the pathology reports (Table \ref{table:fig4}). These variables were selected based on their clinical relevance and alignment with the protocol for examination of radical prostatectomy specimens by the College of American Pathologists \cite{RN17}.

	\begin{table}
	\centering
	\includegraphics[width=0.8\linewidth]{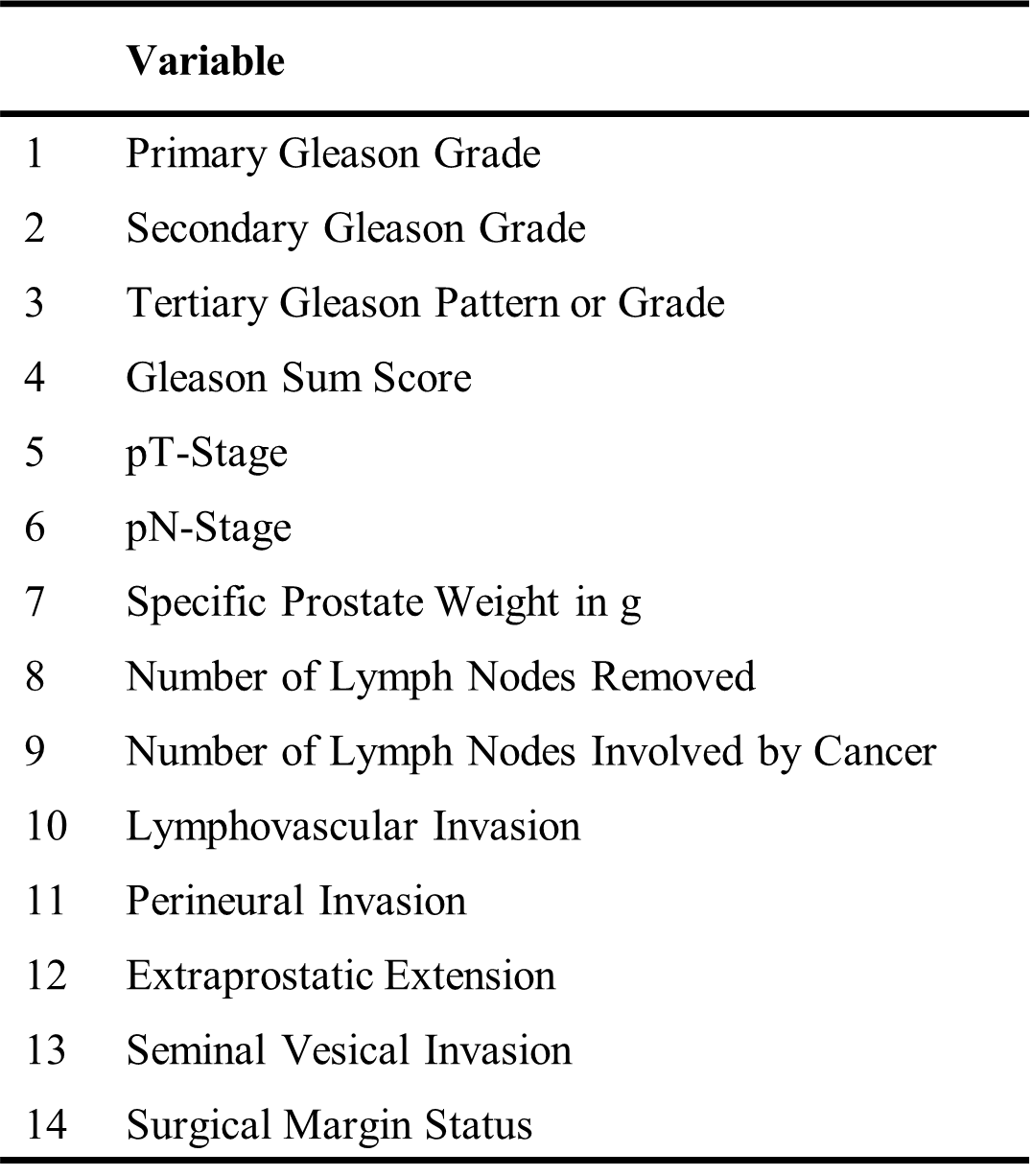}
	\caption{List of abstracted variables.}
	\label{table:fig4}
	\end{table}

	Responses from the three human abstractors were collected using a custom-built website that displayed each pathology report in a PDF viewer, accompanied by an input form designed to capture the abstracted data. Each variable featured a specific set of dropdown answer choices, ensuring that only predefined answers could be selected, thereby precluding the possibility of random or extraneous responses. Responses, along with processing times, were logged for each abstracted report. 
	
	To extract data using the NLP tool, the following prompt was utilized: “Complete the following list of variables with the corresponding values extracted from the given pathology report: pT-Stage, Primary Gleason Grade, Secondary Gleason Grade, Gleason Sum Score, Tertiary Gleason Pattern or Grade, Extraprostatic Extension, Seminal Vesical Invasion, Lymphovascular Invasion, Perineural Invasion, Surgical Margin Status, pN-Stage, Number of Lymph Nodes Removed, Number of Lymph Nodes Involved by Cancer, Specific Prostate Weight in g.” Additionally, the program was instructed to “Compose a comprehensive reply to the query using the search results given. If the search results mention multiple subjects with the same name, create separate answers for each. Only include information found in the results and don't add any additional information. Make sure the answer is correct and don't output false content. If the text does not relate to the query, simply state `Found Nothing'. Ignore outlier search results which has nothing to do with the question. Only answer what is asked. The answer should be short and concise.” 
	
	The mean time required for data abstraction was calculated for the three human abstractors as well as the NLP tool using the two sets of PDF reports. Similarly, the accuracy of data abstraction was calculated for these groups in terms of overall accuracy for the combined set of datapoints (14 variables x 199 reports = 2,786 datapoints) and at the individual variable level. For this analysis, the ground truth (i.e. the correct answer for each data point) was established from the individual responses provided by the human abstractors. The ground truth response was defined when there was agreement between at least two of the data abstractors. In cases with any degree of disagreement among the abstractors, the report was re-examined and if a clear error was identified on part of the dissenting abstractor, the majority consensus was upheld as the ground truth. However, if the re-examination did not reveal an obvious error, further discussions between the abstractors were held until consensus was reached. In instances where agreement between the three abstractors could not be achieved, a fourth reviewer was involved.

	\textit{\textbf{Statistical Analysis}}. Differences in data abstraction times were assessed for statistical significance using the paired Student’s t-test. Comparisons were separately made between the NLP tool and each human abstractor for the two sets of PDF reports. Additionally, the performance of the NLP was compared to itself in a paired fashion using the two datasets. For comparisons of superiority, a P-value of <0.05 was considered statistically significant. A similar paired analysis was performed to assess the accuracy of data abstraction using the McNemar's test of paired proportions. The accuracy of the NLP tool was assessed for non-inferiority to the human abstractors using a non-inferiority margin of -10\% ($\alpha$ = 0.025). Data analysis and visualization was performed using R version 4.0.2 (R Foundation for Statistical Computing).
	
	\section{Results}
\begin{figure}[t]
			\centering
			\includegraphics[width=1\linewidth]{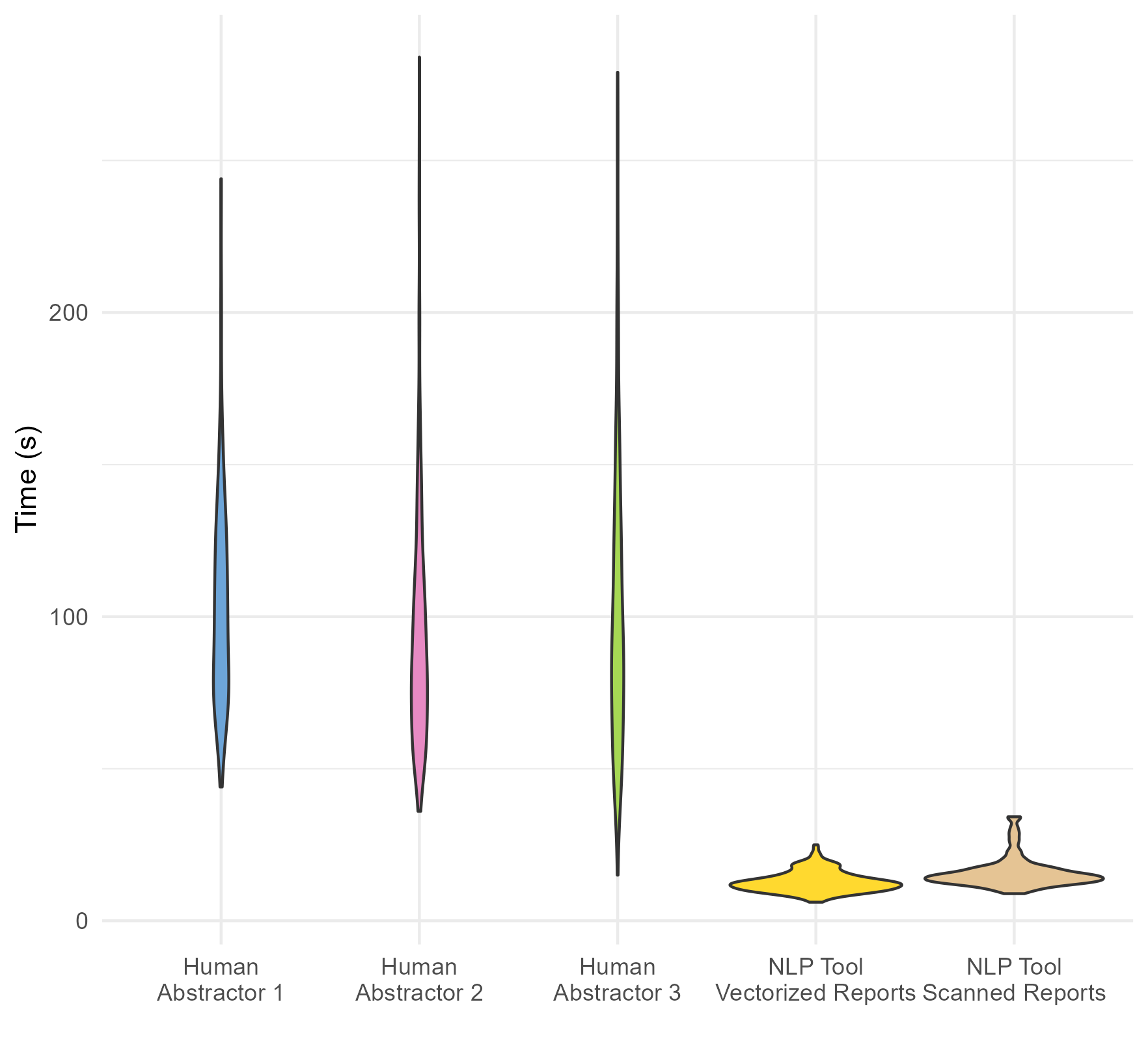}
			\caption{Violin plot showing the distribution of times required for data abstraction.}
			\label{fig:fig2}
		\end{figure}

	As a group, the human abstractors required a mean of 101 seconds (95\% CI, 97 to 104 seconds) per report for data abstraction, with times varying from 15 to 284 seconds. Assessed individually, the three human abstractors had data collection times of 104 seconds (95\% CI 99 to 109 seconds), 94 seconds (95\% CI 89 to 100 seconds), and 103 seconds (95\% CI 96 to 110), respectively.  In comparison, the software tool required a mean of 12.8 second (95\% CI, 12.3 to 13.3 seconds) to process the vectorized reports and a mean of 15.8 seconds (95\% CI 15.1 to 16.5 seconds) to process the scanned reports (P < 0.001 for all paired comparisons). The NLP tool required significantly less time to process the vectorized reports than the scanned documents (P < 0.001). Our observations with respect to data abstraction times are highlighted in  Figure \ref{fig:fig2}.
	
	Compared to the ground truth, the overall accuracies of the three human abstractors were  94.7\% (95\% CI, 93.8 to 95.5\%), 97.8\% (95\% CI, 97.2 to 98.3\%), and 96.4\% (95\% CI, 95.6 to 97\%) for the combined set of 2786 datapoints. In contrast, the software tool achieved an overall accuracy of 94.2\% (95\% CI, 93.3 to 94.9\%) for the vectorized reports and 88.7\% (95\% CI 87.5 to 89.9\%) for the scanned reports. The overall accuracy of the tool for abstracting data from the vectorized reports proved to be non-inferior to each of the three human abstractors (Supplementary Figure \ref{suppfigure:supp2}). For the scanned reports, the tool was non-inferior to 2 out of 3 human abstractors.  A significantly higher degree of accuracy was achieved by the tool for the task of overall data abstraction when using the vectorized reports (Supplementary Figure \ref{suppfigure:supp3}).  
	
		 \begin{figure*}[h!]
		\centering
		\includegraphics[width=1\textwidth]{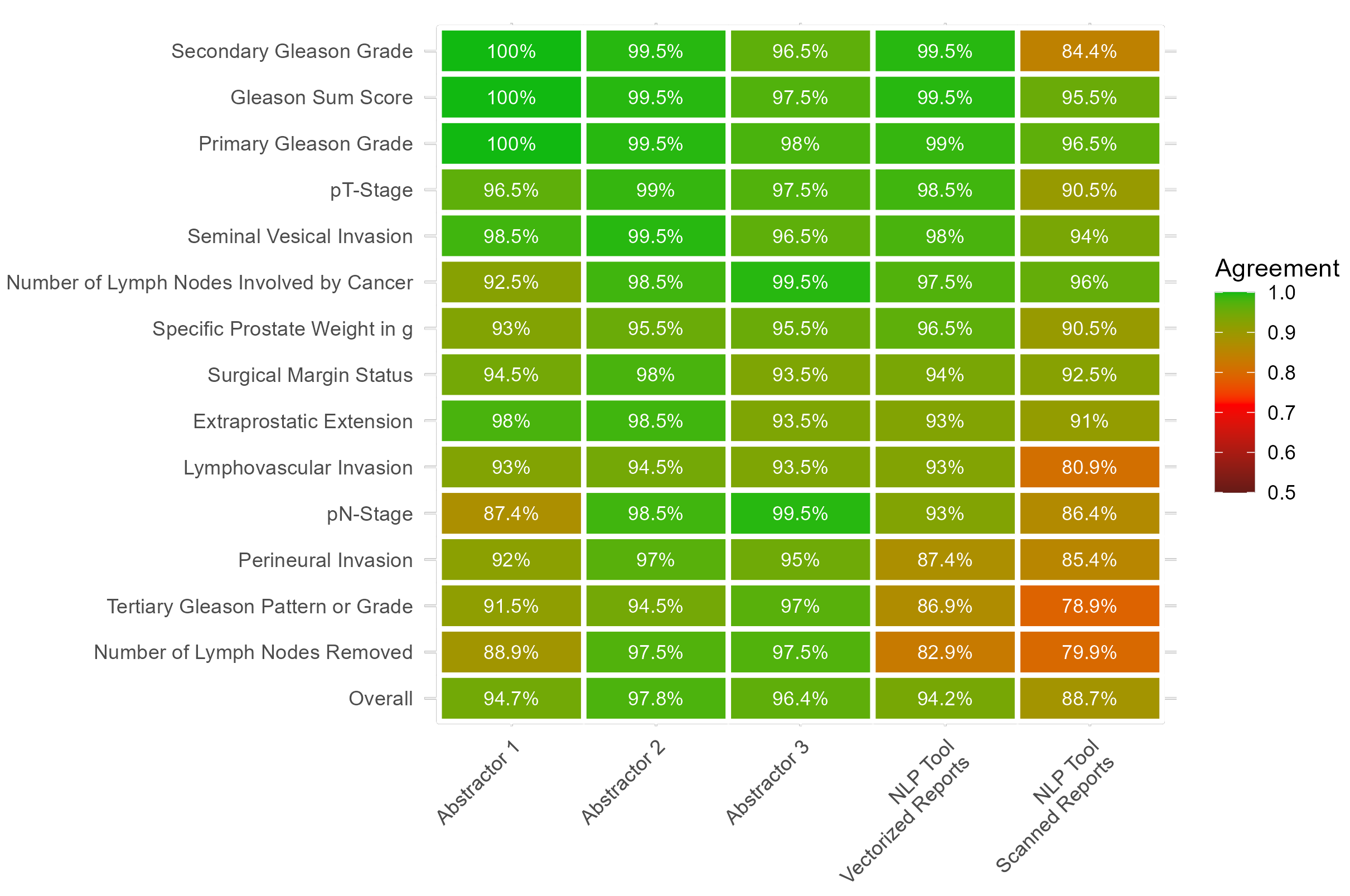}
		\caption{Accuracy of data abstraction by the three humans and NLP tool.}
		\label{fig:fig3}
	\end{figure*}
	
	Two of the three human abstractors achieved an accuracy of >90\% for each of the 14 individual variables. The third abstractor achieved this level of accuracy for 12 out of 14 variables. In contrast, the software achieved >90\% accuracy for 11 variables using the vectorized reports whereas this level of accuracy was achieved for only 8 parameters when using the scanned reports. 
	
	Figure \ref{fig:fig3} provides a summary of the accuracies achieved by the three human abstractors as well as the NLP tool using both datasets. 
	
	\section{Discussion}
	The present study explored the time efficiency and accuracy of a zero-shot learning NLP tool for abstracting data from PDF pathology reports of radical prostatectomy specimens. Our data show that the tool is capable of achieving a 7-fold increase in processing speed compared to human abstractors while maintaining an overall abstraction accuracy of 94\%. Notably, the tool was non-inferior to the three humans at a -10\% margin for the overall task of abstracting 2,786 datapoints from vectorized PDF documents. Although the tool performed slightly worse using the dataset of scanned reports, non-inferiority was achieved relative to two out of the three human abstractors. The performance of the software tool varied across the 14 variables examined, reaching an accuracy of over 90\% accuracy in 11 out of 14 variables. However, the tool struggled with certain variables such as "Tertiary Gleason Pattern" or "Number of Lymph Nodes Removed,” achieving accuracies in the range of 83\%. 
	
	Previous research has employed various machine learning algorithms to parse non-structured data from EHR systems \cite{RN27, RN43, RN41, RN31, RN46, RN40, RN39, RN33, RN6, RN35, RN42, RN48}. However, these tools have uniformly required complex supervised training methods. This requirement is highly labor intensive and limits the number of variables one can practically extract. Furthermore, the reliance on supervised and semi-supervised learning methods limits generalizability of tools to unfamiliar sources of data. These limitations were evident in a study by McCowan et al. \cite{RN46} who used a rule-based and support vector machine classifiers to extract information on  tumor (T) and nodal (N) stages from lung cancer pathology reports.  This study required a training set of 710 reports and yet only achieved accuracy levels of 74\% and 86\% for T and N stage classification, respectively.  Abedian et al. \cite{RN27} looked the broaden the capability of NLP tools for abstracting cancer staging data from a wider range of cancer types, but this required and enormous training data from 555,681 pathology reports and was still only limited to four outputs (ICD10 code and T, N, and metastasis stage). While other more advanced tools, such as DeepPhe, have been specifically engineered to widen the breath of data abstraction capabilities, such tools still require supervised training methods and/or have seen only limited validated in a handful of cancer types \cite{RN33}.
	
	Compared to the tools mentioned above, our method utilizes a zero-shot learning strategy, which requires no task-specific training. This gives our tool the capacity to abstract data on any topic from a body of PDF documents. As urologists, the application described in this report was based on clinical data abstraction from radical prostatectomy reports, but the developed tool could be employed for any number of applications within the field of medicine. Additionally, our tool has the potential be used across a broad range of industries that relay on insights from data contained in free-text format. 
	
	For the development of the software, we specifically utilized the publicly available text-davinci-003 model from OpenAI’s GPT-3 family. Compared to other iterations of GPT-3, the model used in our study was previously shown to excel in tasks like Named Entity Recognition (NER) to identify and categorize key information from unstructured text \cite{RN18}. This study also showed that the  text-davinci-003 model was particularly well suited for interpreting complex sentences and phrases, as shown in tasks like Part-of-speech Tagging \cite{RN37} and Semantic Matching \cite{RN38}.
	
	The prompt crafted by the user and sent to text-davinci-003 model is crucial for achieving the desired output. GPT is highly sensitive to the form and choice of wording, and there are even specialists, known as "prompt engineers", who excel in this task \cite{RN19, RN20}. An example of this can be found in a recent study that evaluated the specific manner in which multiple choice questions are presented to an LLM can significantly impact the accuracy of response \cite{RN21}. These authors found that simply asking multiple choice questions of the model in the usual manner resulted in an accuracy of 55.8\% when indeterminate responses were censored and 36.1\% when indeterminate responses were included on the United States Medical Licensing Examination Step 1. When the users next asked the model to provide a rationale for each answer choice, the accuracy increased to 64.5\% when indeterminate responses were censored and 41.2\% when indeterminate responses were included.
	
	While our model was highly accurate for the abstraction of 11 of the selected variables, more modest results were achieved for the remaining 3. As described by Zheng et al. \cite{RN22}, the majority of incorrect responses from LLMs can be grouped into four categories of error: (1) comprehension, (2) factualness, (3) specificity, and (4) inference. In errors of comprehension, the model does not fully understand the context or the intent of the question or the information in the report, whereas for errors of factualness the model recalls incorrect or outdated information. In terms of specificity, these errors occur because the model does not provide enough information that is relevant to the question to derive a useful answer. For example if it should extract the Gleason score 8, it answers with “high” instead of providing the specific number. Finally, for inference errors, the model might have all the necessary information to answer a question, but it fails to “reason” with the facts effectively to arrive at the correct answer.
	
	The tool's performance is not solely dependent on the GPT component or prompt generation. In the application selected for our tool, parsing PDFs is a critical aspect of the process. Various Python libraries are available for PDF parsing, but their efficacy can vary, potentially leading to data loss or errors during the conversion from PDF to text. Particularly, data in complex layouts or tabular formats might be misinterpreted. We found in our sub-analysis of the scanned reports, that OCR made a difference in overall accuracy of 5.5\%, leading to a decrease in performance of the NLP tool to 88.7\%. 
	
	Despite the challenges mentioned, we feel the developed tool represents a significant advancement in the ability for researchers to abstract data without the need for individual NLP model creation or training. We feel that if vectorized PDFs or improved OCR and text extraction methods are utilized with carefully crafted prompts specific to individual variables, the error rate achieved would balance well against the gain of 7-fold improvement in abstraction efficiency, freeing human abstractors to perform more critical research tasks. We should note, however, that the tool developed for this research report is not quite ready for immediate use in clinical practice, as it does not include end-to-end encryption which is required to meet the standards of the United States Health Insurance Portability and Accountability Act (HIPAA) or the General Data Protection Regulation (GDPR) of the European Union.

	\section{Conclusion}
	The presented zero-shot learning NLP tool affords researchers comparable levels of accuracy to that of human abstractors, with significant time savings benefits. Because our tool does not require task-specific training, it can be adopted for a limitless number of data abstraction tasks, including applications outside the field of medicine. Continued refinement, particularly in terms of optical character recognition (OCR) capabilities, may enhance its practical use, speed, and degree of accuracy. 

\clearpage

	\section{Acknowledgements}
	The research was supported by a grant from the Swiss Cancer League (Krebsliga Schweiz), Swiss Urological Society, and the Menon Family Foundation.  

	\normalsize
	\bibliography{references}

\begin{thebibliography}{35}
\providecommand{\natexlab}[1]{#1}
\providecommand{\url}[1]{\texttt{#1}}
\expandafter\ifx\csname urlstyle\endcsname\relax
  \providecommand{\doi}[1]{doi: #1}\else
  \providecommand{\doi}{doi: \begingroup \urlstyle{rm}\Url}\fi

\bibitem[Garza et~al.(2022)Garza, Williams, Myneni, Fenton, Ounpraseuth, Hu,
  Lee, Snowden, Zozus, and Walden]{RN2}
Maryam~Y Garza, Tremaine Williams, Sahiti Myneni, Susan~H Fenton, Songthip
  Ounpraseuth, Zhuopei Hu, Jeannette Lee, Jessica Snowden, Meredith~N Zozus,
  and Anita~C Walden.
\newblock Measuring and controlling medical record abstraction (mra) error
  rates in an observational study.
\newblock \emph{BMC Medical Research Methodology}, 22\penalty0 (1):\penalty0
  227, 2022.
\newblock ISSN 1471-2288.

\bibitem[Kong(2019)]{RN3}
Hyoun-Joong Kong.
\newblock Managing unstructured big data in healthcare system.
\newblock \emph{Healthcare informatics research}, 25\penalty0 (1):\penalty0
  1--2, 2019.

\bibitem[Polnaszek et~al.(2016)Polnaszek, Gilmore-Bykovskyi, Hovanes, Roiland,
  Ferguson, Brown, and Kind]{RN1}
Brock Polnaszek, Andrea Gilmore-Bykovskyi, Melissa Hovanes, Rachel Roiland,
  Patrick Ferguson, Roger Brown, and Amy~JH Kind.
\newblock Overcoming the challenges of unstructured data in multi-site,
  electronic medical record-based abstraction.
\newblock \emph{Medical care}, 54\penalty0 (10):\penalty0 e65, 2016.

\bibitem[Alzoubi et~al.(2019)Alzoubi, Alzubi, Ramzan, West, Al-Hadhrami, and
  Alazab]{RN5}
Hadeel Alzoubi, Raid Alzubi, Naeem Ramzan, Daune West, Tawfik Al-Hadhrami, and
  Mamoun Alazab.
\newblock A review of automatic phenotyping approaches using electronic health
  records.
\newblock \emph{Electronics}, 8\penalty0 (11):\penalty0 1235, 2019.
\newblock ISSN 2079-9292.

\bibitem[Miller et~al.(2022)Miller, Fafaj, Tastaldi, Alkhatib, Zolin,
  AlMarzooqi, Tu, Alaedeen, Prabhu, and Krpata]{RN7}
Benjamin~T Miller, Aldo Fafaj, Luciano Tastaldi, Hemasat Alkhatib, Samuel
  Zolin, Raha AlMarzooqi, Chao Tu, Diya Alaedeen, Ajita~S Prabhu, and David~M
  Krpata.
\newblock Capturing surgical data: Comparing a quality improvement registry to
  natural language processing and manual chart review.
\newblock \emph{Journal of Gastrointestinal Surgery}, 26\penalty0 (7):\penalty0
  1490--1494, 2022.
\newblock ISSN 1091-255X.

\bibitem[Savova et~al.(2019)Savova, Danciu, Alamudun, Miller, Lin, Bitterman,
  Tourassi, and Warner]{RN6}
Guergana~K Savova, Ioana Danciu, Folami Alamudun, Timothy Miller, Chen Lin,
  Danielle~S Bitterman, Georgia Tourassi, and Jeremy~L Warner.
\newblock Use of natural language processing to extract clinical cancer
  phenotypes from electronic medical records.
\newblock \emph{Cancer research}, 79\penalty0 (21):\penalty0 5463--5470, 2019.
\newblock ISSN 0008-5472.

\bibitem[Sheikhalishahi et~al.(2019)Sheikhalishahi, Miotto, Dudley, Lavelli,
  Rinaldi, and Osmani]{RN4}
Seyedmostafa Sheikhalishahi, Riccardo Miotto, Joel~T Dudley, Alberto Lavelli,
  Fabio Rinaldi, and Venet Osmani.
\newblock Natural language processing of clinical notes on chronic diseases:
  systematic review.
\newblock \emph{JMIR medical informatics}, 7\penalty0 (2):\penalty0 e12239,
  2019.

\bibitem[Dunn et~al.(2022)Dunn, Dagdelen, Walker, Lee, Rosen, Ceder, Persson,
  and Jain]{RN24}
Alexander Dunn, John Dagdelen, Nicholas Walker, Sanghoon Lee, Andrew~S Rosen,
  Gerbrand Ceder, Kristin Persson, and Anubhav Jain.
\newblock Structured information extraction from complex scientific text with
  fine-tuned large language models.
\newblock \emph{arXiv preprint arXiv:2212.05238}, 2022.

\bibitem[Leyh-Bannurah et~al.(2018)Leyh-Bannurah, Tian, Karakiewicz, Wolffgang,
  Sauter, Fisch, Pehrke, Huland, Graefen, and Budäus]{RN31}
Sami-Ramzi Leyh-Bannurah, Zhe Tian, Pierre~I Karakiewicz, Ulrich Wolffgang,
  Guido Sauter, Margit Fisch, Dirk Pehrke, Hartwig Huland, Markus Graefen, and
  Lars Budäus.
\newblock Deep learning for natural language processing in urology:
  State-of-the-art automated extraction of detailed pathologic prostate cancer
  data from narratively written electronic health records.
\newblock \emph{JCO clinical cancer informatics}, 2:\penalty0 1--9, 2018.
\newblock ISSN 2473-4276.

\bibitem[Romera-Paredes and Torr()]{RN8}
Bernardino Romera-Paredes and Philip Torr.
\newblock An embarrassingly simple approach to zero-shot learning.
\newblock In \emph{International conference on machine learning}, pages
  2152--2161. PMLR.

\bibitem[Sun et~al.(2021)Sun, Gu, and Sun]{RN9}
Xiaohong Sun, Jinan Gu, and Hongying Sun.
\newblock Research progress of zero-shot learning.
\newblock \emph{Applied Intelligence}, 51:\penalty0 3600--3614, 2021.
\newblock ISSN 0924-669X.

\bibitem[Yim et~al.(2016)Yim, Yetisgen, Harris, and Kwan]{RN29}
Wen-wai Yim, Meliha Yetisgen, William~P Harris, and Sharon~W Kwan.
\newblock Natural language processing in oncology: a review.
\newblock \emph{JAMA oncology}, 2\penalty0 (6):\penalty0 797--804, 2016.
\newblock ISSN 2374-2437.

\bibitem[RN1({\natexlab{a}})]{RN11}
Pymupdf. version 1.22.3. retrieved from https://github.com/pymupdf/pymupdf,
  {\natexlab{a}}.

\bibitem[Cer et~al.(2018)Cer, Yang, Kong, Hua, Limtiaco, John, Constant,
  Guajardo-Cespedes, Yuan, and Tar]{RN12}
Daniel Cer, Yinfei Yang, Sheng-yi Kong, Nan Hua, Nicole Limtiaco, Rhomni~St
  John, Noah Constant, Mario Guajardo-Cespedes, Steve Yuan, and Chris Tar.
\newblock Universal sentence encoder.
\newblock \emph{arXiv preprint arXiv:1803.11175}, 2018.

\bibitem[RN1({\natexlab{b}})]{RN13}
Scikit-learn. version 1.2.2. retrieved from https://scikit-learn.org/stable/.
\newblock {\natexlab{b}}.

\bibitem[RN1({\natexlab{c}})]{RN14}
Openai python library. version 0.27.7. retrieved from
  https://pypi.org/project/openai/.
\newblock {\natexlab{c}}.

\bibitem[Ye et~al.(2023)Ye, Chen, Xu, Zu, Shao, Liu, Cui, Zhou, Gong, and
  Shen]{RN18}
Junjie Ye, Xuanting Chen, Nuo Xu, Can Zu, Zekai Shao, Shichun Liu, Yuhan Cui,
  Zeyang Zhou, Chao Gong, and Yang Shen.
\newblock A comprehensive capability analysis of gpt-3 and gpt-3.5 series
  models.
\newblock \emph{arXiv preprint arXiv:2303.10420}, 2023.

\bibitem[Institute()]{RN16}
National~Cancer Institute.
\newblock The cancer genome atlas program (tcga).

\bibitem[Paner et~al.(2021)Paner, Srigley, Pettus, Giannico, Sirintrapun, and
  Harik]{RN17}
G~Paner, J~Srigley, Jason Pettus, Giovanna~Angela Giannico, Joseph Sirintrapun,
  and Lara~R Harik.
\newblock Protocol for the examination of radical prostatectomy specimens from
  patients with carcinoma of the prostate gland.
\newblock \emph{College of American Pathologists (CAP): Northfield, IL, USA},
  2021.

\bibitem[Abedian et~al.(2021)Abedian, Sholle, Adekkanattu, Cusick, Weiner,
  Shoag, Hu, and Campion~Jr]{RN27}
Sajjad Abedian, Evan~T Sholle, Prakash~M Adekkanattu, Marika~M Cusick,
  Stephanie~E Weiner, Jonathan~E Shoag, Jim~C Hu, and Thomas~R Campion~Jr.
\newblock Automated extraction of tumor staging and diagnosis information from
  surgical pathology reports.
\newblock \emph{JCO Clinical Cancer Informatics}, 5:\penalty0 1054--1061, 2021.
\newblock ISSN 2473-4276.

\bibitem[Glaser et~al.(2018)Glaser, Jordan, Cohen, Desai, Silberman, and
  Meeks]{RN43}
Alexander~P Glaser, Brian~J Jordan, Jason Cohen, Anuj Desai, Philip Silberman,
  and Joshua~J Meeks.
\newblock Automated extraction of grade, stage, and quality information from
  transurethral resection of bladder tumor pathology reports using natural
  language processing.
\newblock \emph{JCO Clinical Cancer Informatics}, 2:\penalty0 1--8, 2018.
\newblock ISSN 2473-4276.

\bibitem[Kim et~al.(2014)Kim, Merchant, Zheng, Thomas, Contreras, Jacobsen, and
  Chien]{RN41}
B.~J. Kim, M.~Merchant, C.~Zheng, A.~A. Thomas, R.~Contreras, S.~J. Jacobsen,
  and G.~W. Chien.
\newblock A natural language processing program effectively extracts key
  pathologic findings from radical prostatectomy reports.
\newblock \emph{J Endourol}, 28\penalty0 (12):\penalty0 1474--8, 2014.
\newblock ISSN 0892-7790.
\newblock \doi{10.1089/end.2014.0221}.

\bibitem[McCowan et~al.(2007)McCowan, Moore, Nguyen, Bowman, Clarke, Duhig, and
  Fry]{RN46}
Iain~A McCowan, Darren~C Moore, Anthony~N Nguyen, Rayleen~V Bowman, Belinda~E
  Clarke, Edwina~E Duhig, and Mary-Jane Fry.
\newblock Collection of cancer stage data by classifying free-text medical
  reports.
\newblock \emph{Journal of the American Medical Informatics Association},
  14\penalty0 (6):\penalty0 736--745, 2007.
\newblock ISSN 1527-974X.

\bibitem[McCowan et~al.()McCowan, Moore, and Fry]{RN40}
Iain McCowan, Darren Moore, and Mary-Jane Fry.
\newblock Classification of cancer stage from free-text histology reports.
\newblock In \emph{2006 International Conference of the IEEE Engineering in
  Medicine and Biology Society}, pages 5153--5156. IEEE.
\newblock ISBN 1424400325.

\bibitem[Nguyen et~al.(2010)Nguyen, Lawley, Hansen, Bowman, Clarke, Duhig, and
  Colquist]{RN39}
Anthony~N Nguyen, Michael~J Lawley, David~P Hansen, Rayleen~V Bowman, Belinda~E
  Clarke, Edwina~E Duhig, and Shoni Colquist.
\newblock Symbolic rule-based classification of lung cancer stages from
  free-text pathology reports.
\newblock \emph{Journal of the American Medical Informatics Association},
  17\penalty0 (4):\penalty0 440--445, 2010.
\newblock ISSN 1527-974X.

\bibitem[Savova et~al.(2017)Savova, Tseytlin, Finan, Castine, Miller,
  Medvedeva, Harris, Hochheiser, Lin, Chavan, and Jacobson]{RN33}
G.~K. Savova, E.~Tseytlin, S.~Finan, M.~Castine, T.~Miller, O.~Medvedeva,
  D.~Harris, H.~Hochheiser, C.~Lin, G.~Chavan, and R.~S. Jacobson.
\newblock Deepphe: A natural language processing system for extracting cancer
  phenotypes from clinical records.
\newblock \emph{Cancer Res}, 77\penalty0 (21):\penalty0 e115--e118, 2017.
\newblock ISSN 0008-5472 (Print) 0008-5472.
\newblock \doi{10.1158/0008-5472.Can-17-0615}.

\bibitem[Warner et~al.(2016)Warner, Levy, Neuss, Warner, Levy, and Neuss]{RN35}
Jeremy~L Warner, Mia~A Levy, Michael~N Neuss, Jeremy~L Warner, Mia~A Levy, and
  Michael~N Neuss.
\newblock Recap: feasibility and accuracy of extracting cancer stage
  information from narrative electronic health record data.
\newblock \emph{Journal of oncology practice}, 12\penalty0 (2):\penalty0
  157--158, 2016.
\newblock ISSN 1554-7477.

\bibitem[Yala et~al.(2017)Yala, Barzilay, Salama, Griffin, Sollender, Bardia,
  Lehman, Buckley, Coopey, and Polubriaginof]{RN42}
Adam Yala, Regina Barzilay, Laura Salama, Molly Griffin, Grace Sollender,
  Aditya Bardia, Constance Lehman, Julliette~M Buckley, Suzanne~B Coopey, and
  Fernanda Polubriaginof.
\newblock Using machine learning to parse breast pathology reports.
\newblock \emph{Breast cancer research and treatment}, 161:\penalty0 203--211,
  2017.
\newblock ISSN 0167-6806.

\bibitem[Zhou et~al.(2022)Zhou, Wang, Wang, Liu, and Zhang]{RN48}
Sicheng Zhou, Nan Wang, Liwei Wang, Hongfang Liu, and Rui Zhang.
\newblock Cancerbert: a cancer domain-specific language model for extracting
  breast cancer phenotypes from electronic health records.
\newblock \emph{Journal of the American Medical Informatics Association},
  29\penalty0 (7):\penalty0 1208--1216, 2022.
\newblock ISSN 1527-974X.

\bibitem[Schmid(1994)]{RN37}
Helmut Schmid.
\newblock Part-of-speech tagging with neural networks.
\newblock \emph{arXiv preprint cmp-lg/9410018}, 1994.

\bibitem[Jiang et~al.()Jiang, Zhang, Li, Bendersky, Golbandi, and Najork]{RN38}
Jyun-Yu Jiang, Mingyang Zhang, Cheng Li, Michael Bendersky, Nadav Golbandi, and
  Marc Najork.
\newblock Semantic text matching for long-form documents.
\newblock In \emph{The world wide web conference}, pages 795--806.

\bibitem[Lee et~al.(2023)Lee, Bubeck, and Petro]{RN19}
Peter Lee, Sebastien Bubeck, and Joseph Petro.
\newblock Benefits, limits, and risks of gpt-4 as an ai chatbot for medicine.
\newblock \emph{New England Journal of Medicine}, 388\penalty0 (13):\penalty0
  1233--1239, 2023.
\newblock ISSN 0028-4793.

\bibitem[White et~al.(2023)White, Fu, Hays, Sandborn, Olea, Gilbert, Elnashar,
  Spencer-Smith, and Schmidt]{RN20}
Jules White, Quchen Fu, Sam Hays, Michael Sandborn, Carlos Olea, Henry Gilbert,
  Ashraf Elnashar, Jesse Spencer-Smith, and Douglas~C Schmidt.
\newblock A prompt pattern catalog to enhance prompt engineering with chatgpt.
\newblock \emph{arXiv preprint arXiv:2302.11382}, 2023.

\bibitem[Kung et~al.(2023)Kung, Cheatham, Medenilla, Sillos, De~Leon, Elepaño,
  Madriaga, Aggabao, Diaz-Candido, and Maningo]{RN21}
Tiffany~H Kung, Morgan Cheatham, Arielle Medenilla, Czarina Sillos, Lorie
  De~Leon, Camille Elepaño, Maria Madriaga, Rimel Aggabao, Giezel
  Diaz-Candido, and James Maningo.
\newblock Performance of chatgpt on usmle: Potential for ai-assisted medical
  education using large language models.
\newblock \emph{PLoS digital health}, 2\penalty0 (2):\penalty0 e0000198, 2023.
\newblock ISSN 2767-3170.

\bibitem[Zheng et~al.(2023)Zheng, Huang, and Chang]{RN22}
Shen Zheng, Jie Huang, and Kevin Chen-Chuan Chang.
\newblock Why does chatgpt fall short in answering questions faithfully?
\newblock \emph{arXiv preprint arXiv:2304.10513}, 2023.

\end{thebibliography}
	
\clearpage
\onecolumn
\section{Supplementary Material}
\begin{suppfigure*}[!h]
	\centering
	\includegraphics[width=1\textwidth]{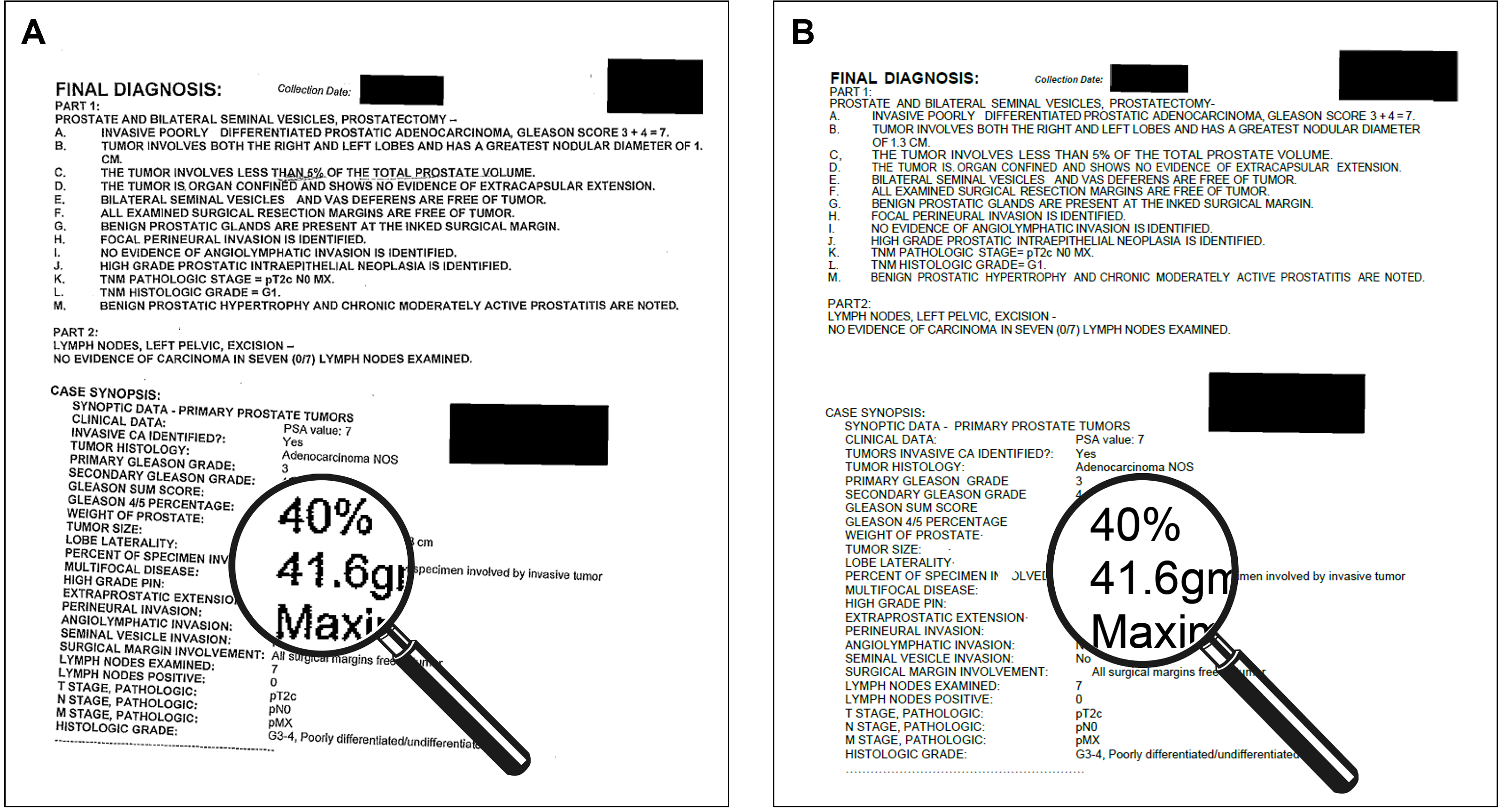}
	\vspace*{-3mm}
	\caption{Example of a scanned (A) and vectorized (B) de-identified radical prostatectomy pathology reports.}\label{suppfigure:supp1}
\end{suppfigure*}

\clearpage

\begin{suppfigure*}[!h]
	\centering
	\includegraphics[width=1\textwidth]{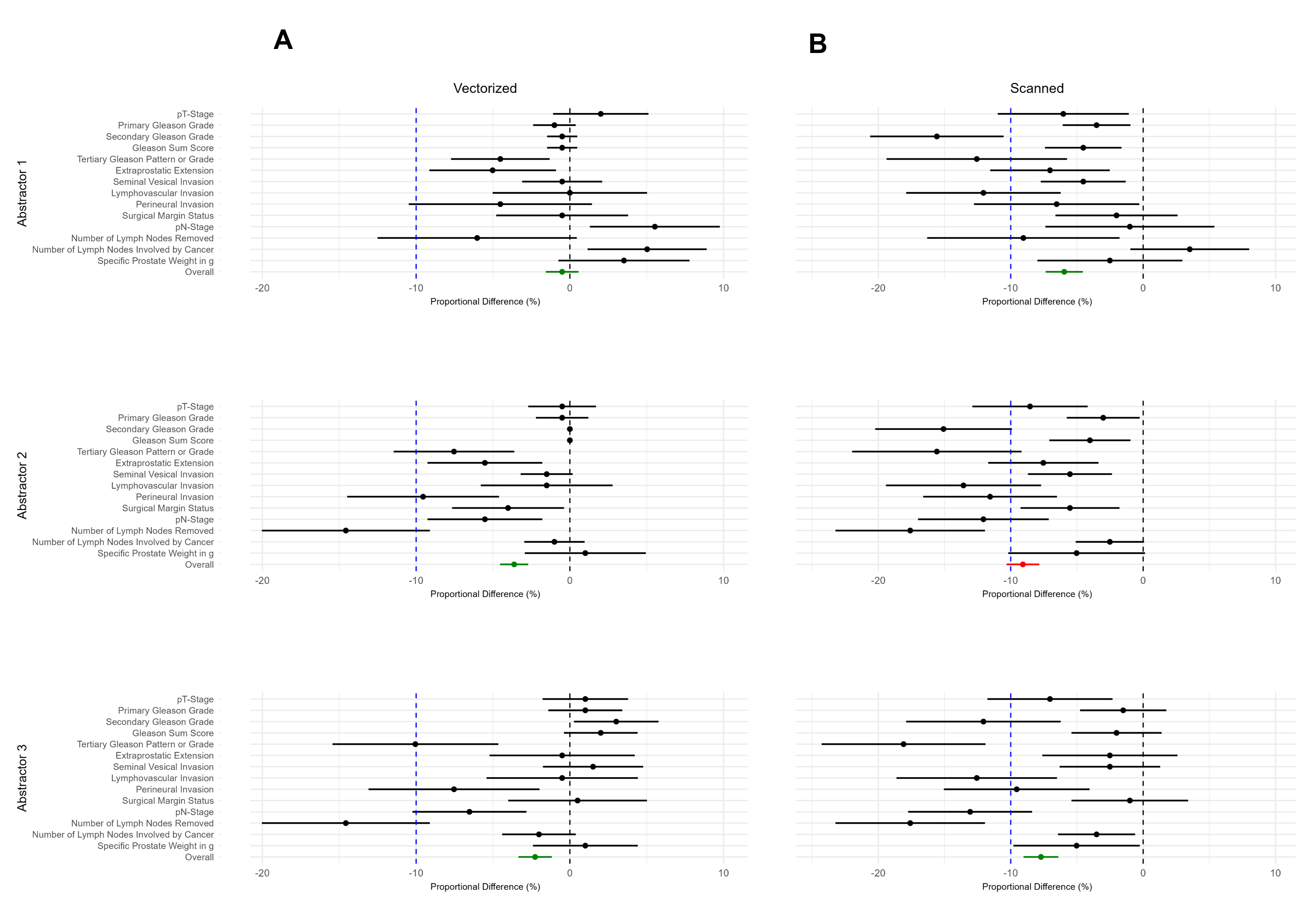}
	\caption{Non-inferiority analysis of accuracy of the NLP tool versus human abstractors using vectorized (A) and scanned reports (B). Lines represent 97.5\% confidence intervals of differences in proportions of accuracy, with points marking the mean difference. The blue dashed line denotes the -10\% non-inferiority margin. }
	\label{suppfigure:supp2}
\end{suppfigure*}
\clearpage

\begin{suppfigure*}[!h]
	\centering
	\includegraphics[width=1\textwidth]{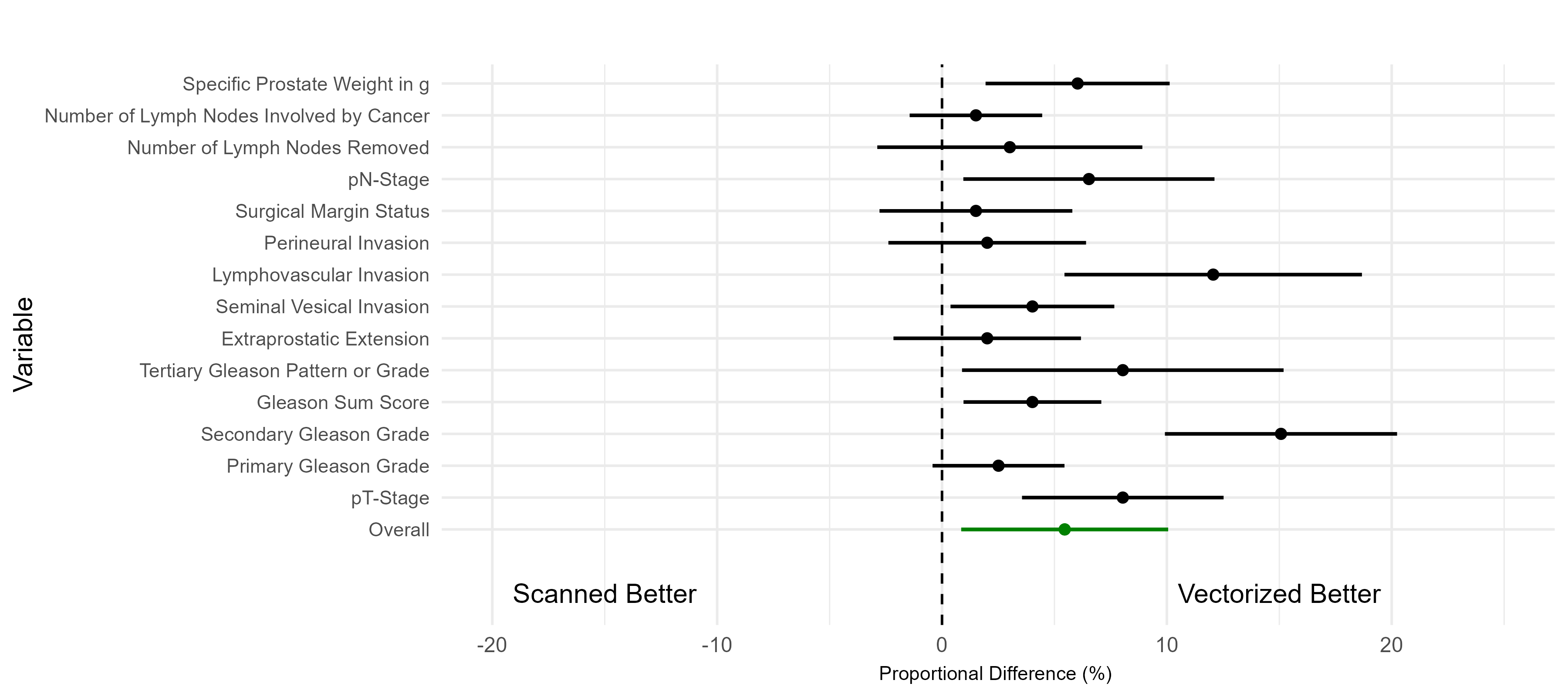}
	\caption{Non-inferiority analysis of accuracy of the NLP tool using vectorized versus scanned reports. Lines represent 97.5\% confidence intervals of differences in proportions of accuracy, with points marking the mean difference.}
	\label{suppfigure:supp3}
\end{suppfigure*}

\end{document}